\documentclass[conference,10pt]{IEEEtran}
\usepackage{algpseudocode}                                 
\usepackage{algorithm}
\usepackage{graphicx}                                      
\usepackage{amsmath}
\usepackage{amssymb}
\usepackage{amsfonts}
\usepackage{amsthm}
\usepackage[mathcal]{eucal}
\usepackage{booktabs}
\usepackage{multirow}
\usepackage[subrefformat=parens,farskip=0pt,justification=centering]{subfig}
\usepackage{color}
\usepackage{cite}                                          
\usepackage{comment}                                       
\usepackage{soul}                                          
\soulregister\cite7
\soulregister\ref7
\soulregister\pageref7
\usepackage{etoolbox}                                      
\usepackage{url}
\usepackage{nth}                                           
\usepackage{bm}                                            
\usepackage{courier}
\usepackage{balance}
\usepackage{threeparttable}
\usepackage[bookmarks=false]{hyperref}
\hypersetup{
    colorlinks = true,
    citecolor  = blue,
    linkcolor  = blue,
    urlcolor   = blue,
}
\usepackage{tikz}
\usetikzlibrary{positioning, fit}
\usepackage{filecontents}                                  
\usepackage{pgfplots}
\usepackage{pgfplotstable}
\pgfplotsset{compat=newest}
\usepackage{caption}
\usepackage{cleveref}
\Crefformat{figure}{Fig.~#2#1#3}                           
\Crefname{subfigure}{Fig.}{Figs.}
\Crefname{figure}{Fig.}{Figs.}
\Crefformat{table}{TABLE~#2#1#3}                           
\definecolor{CUHKorange}{RGB}{244,106,18} 
\definecolor{CUHKblue}{RGB}{0,111,190}    
\definecolor{CUHKgreen}{RGB}{0,127,128}   
\definecolor{CUHKred}{RGB}{228,46,36}     
\definecolor{CUHKyellow}{RGB}{198,148,34} 
\definecolor{CUHKdark}{RGB}{114,44,114}   
\definecolor{CUHKmiddle}{RGB}{144,44,144} 
\definecolor{CUHKlight}{RGB}{167,44,167} 

\renewcommand{\vec}[1]{\boldsymbol{#1}}

\newtheorem{mydefinition}{\textbf{Definition}}

\algrenewcommand\textproc{\texttt}

\makeatletter
\let\OldStatex\Statex
\renewcommand{\Statex}[1][3]{%
  \setlength\@tempdima{\algorithmicindent}%
  \OldStatex\hskip\dimexpr#1\@tempdima\relax
}
\makeatother

\RequirePackage[normalem]{ulem} 
\RequirePackage{color}\definecolor{RED}{rgb}{1,0,0}\definecolor{BLUE}{rgb}{0,0,1} 

\graphicspath{{./figs/}}

\usepackage{flexisym}

\begin{document}

\title{
    Automatic Layout Generation with Applications in Machine Learning Engine Evaluation
}


\author{
    \IEEEauthorblockN{
        Haoyu Yang\IEEEauthorrefmark{1},
        Wen Chen\IEEEauthorrefmark{1},
        Piyush Pathak\IEEEauthorrefmark{2},
        Frank Gennari\IEEEauthorrefmark{2},
        Ya-Chieh Lai\IEEEauthorrefmark{2},
        Bei Yu\IEEEauthorrefmark{1}
    }
    \IEEEauthorblockA{
        \IEEEauthorrefmark{1}The Chinese University of Hong Kong \\
        \IEEEauthorrefmark{2}Cadence Design Systems, Inc.\\
        Email: \normalsize\texttt{\{hyyang,byu\}@cse.cuhk.edu.hk}
    }
}

\maketitle

\begin{abstract}
Machine learning-based lithography hotspot detection has been deeply studied recently,
from varies feature extraction techniques to efficient learning models.
It has been observed that such machine learning-based frameworks are providing satisfactory metal layer hotspot prediction results on known public metal layer benchmarks.
In this work, we seek to evaluate how these machine learning-based hotspot detectors generalize to complicated patterns.
We first introduce a automatic layout generation tool that can synthesize varies layout patterns given a set of design rules.
The tool currently supports both metal layer and via layer generation.
As a case study, we conduct hotspot detection on the generated via layer layouts with representative machine learning-based hotspot detectors,
which shows that continuous study on model robustness and generality is necessary to prototype and integrate the learning engines in DFM flows.
The source code of the layout generation tool will be available at
\url{https://github.com/phdyang007/layout-generation}. 
\end{abstract}

\section{Introduction}

The rapid development of machine learning techniques have brought promising alternatives to design for manufacturability problems, especially for lithography hotspot detection.
Related researches range from varies feature extraction techniques
\cite{HSD-ICCAD2016-Zhang,HSD-ASPDAC2019-Yang,HSD-SPIE2015-Matsunawa,HSD-TCAD2017-Yang,HSD-ASPDAC2017-Tomioka,HSD-ISVLSI2018-Geng,HSD-SOCC2017-Yang}
and efficient learning models \cite{HSD-ICCAD2016-Zhang,HSD-TCAD2019-Yang,HSD-DAC2019-Chen,HSD-ASPDAC2019-Ye,HSD-ASPDAC2012-Ding,HSD-JM3-2017-Yang,HSD-JM3-2016-Shin}.
Most of these frameworks are presenting very high hotspot detection accuracy with reasonable false positive overhead.
However, most of them are targeting at a public benchmark at legacy technology node from ICCAD2012 CAD Contest \cite{HSD-ICCAD2012-Torres},
with carefully designed and tuned parameters and machine learning engines.

In this paper, we seek to evaluate how these machine learning-based hotspot detectors generalize to complicated patterns.
We first introduce an automatic layout generation tool that can synthesize various layout patterns given a set of design rules.
{Unlike the existing pattern generation solutions that seek to increase the pattern space within a constrained Euclidean distance \cite{HSD-DAC2019-DeePattern,HSD-JM3-2015-Shim}, the proposed automatic generation flow generates patterns for certain DRC constrained designs.
}

\begin{figure}[tb!]
	\centering
	\subfloat[Low via density]{\includegraphics[width=.426\linewidth]{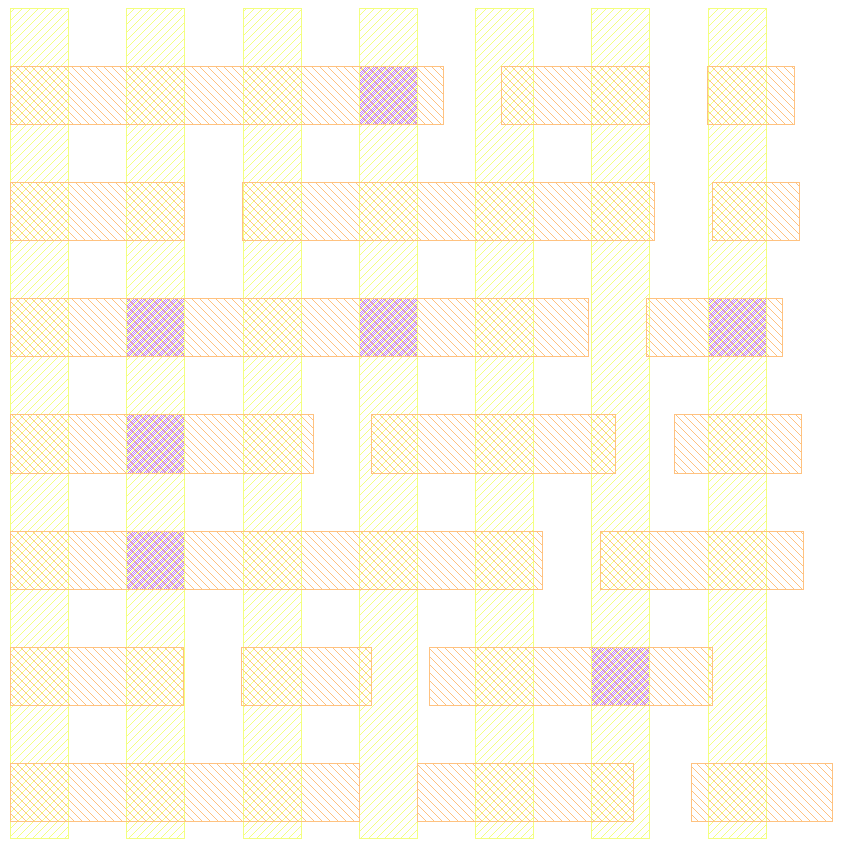}}
    \hspace{.1in}
	\subfloat[High via density]{\includegraphics[width=.426\linewidth]{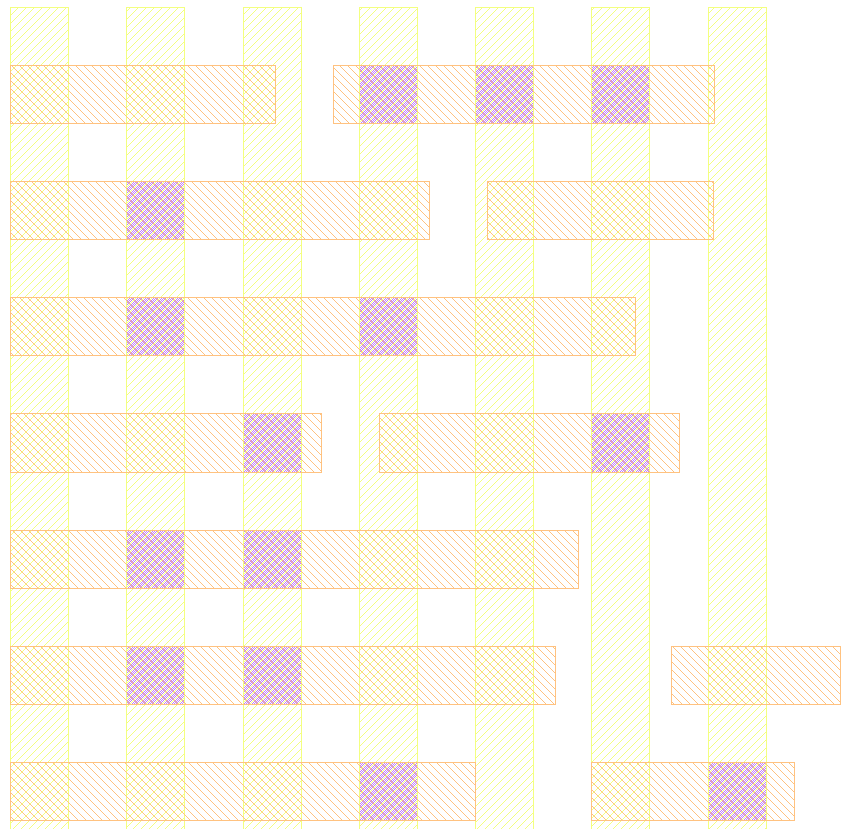}}
	\caption{Visualization of generated via patterns. Vias are randomly placed at the intersection region of given upper and lower metal shapes with a predetermined probability that controls via density.}
	\label{fig:vias}
\end{figure}

The tool supports both metal layer and via layer generation.
In the configuration files, we can manually achieve certain layout properties including tip to tip distance, wire CD, layout density, via density and so on.
These configurations enable a generation of diverse layout patterns under a given technology node,
which can be applied to various DFM research and analysis including hotspot detection and OPC recipe development.
An examples of via layers with different density can be found in \Cref{fig:vias}, where the upper and the lower metal gratings are intentionally placed to assist via generation.

As a case study, we conduct hotspot detection on the generated via layer layouts with representative machine learning-based hotspot detectors \cite{HSD-ICCAD2016-Zhang},\cite{HSD-TCAD2019-Yang},\cite{HSD-ASPDAC2019-Ye},
which adopt smooth-boosting \cite{HSD-ICCAD2016-Zhang}, shallow convolutional neural networks \cite{HSD-TCAD2019-Yang,HSD-ASPDAC2019-Ye}, respectively.
Although these hotspot detectors achieve promising results on ICCAD2012 CAD contest benchmarks, we show performance degradation occurs with complex dataset,
which implies continuous study on model robustness and generality is necessary to prototype and integrate the learning engines in DFM flows.
The contributions of this paper are listed below.
\begin{itemize}
    \item We develop a layout generation toolkit with $\mathsf{Python}$ and $\mathsf{KLayout}$ \cite{KLAYOUT} backbones, 
        to support efficient unidirectional metal layer and via/contact generation.
    \item We feed the generated via layers into a commercial lithography simulator and construct a via layer hotspot benchmark suit for hotspot detector evaluation.
    \item Evaluation experiments show that representative hotspot detectors may fail to exhibit a satisfactory detection performance on either accuracy or false positive penalty.
\end{itemize}

The rest of the paper is organized as follows.
\Cref{sec:tool} describes details of the layout generation flow.
\Cref{sec:exp} presents a case study of state-of-the-art hotspot detector evaluation, followed by conclusion in \Cref{sec:conclu}.

\section{A Layout Generation Toolkit}
\label{sec:tool}

In this section, we will introduce the details of our layout generation flow with $\mathsf{Python}$ and $\mathsf{KLayout}$ backbones.
In current version, we support unidirectional metal layer and via/contact generation.

\subsection{Metal Generation}
\begin{figure}[tb!]
	\centering
	\includegraphics[width=.4\textwidth]{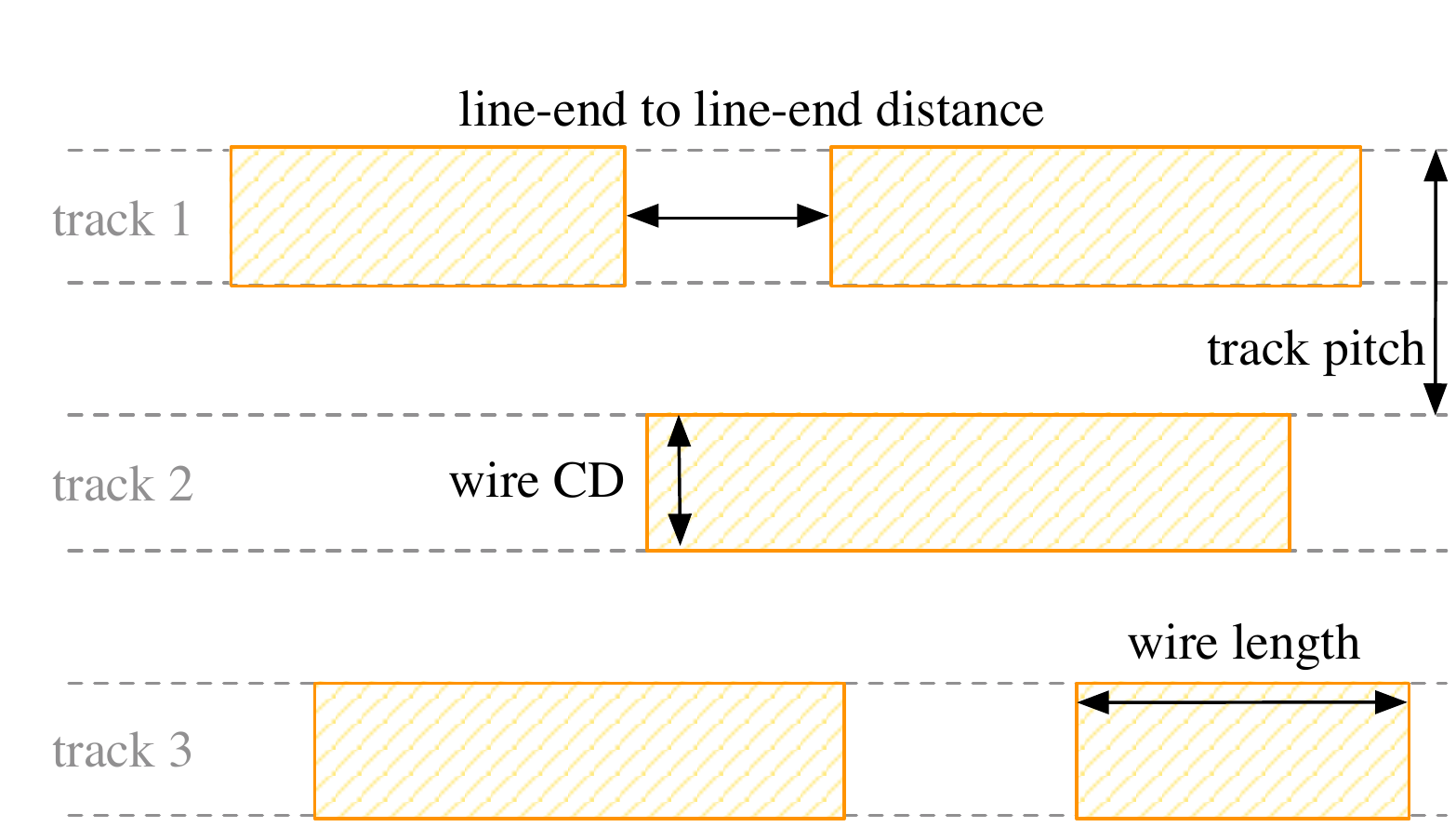}
	\caption{Visualization of terminologies in metal generation constraints.}
	\label{fig:def-metal}
\end{figure}

The layout generation tool generates metal layers cell by cell where metal shape distribution and placement are controlled by a set of design constraints as listed below.
\begin{itemize}
	\item[-] \texttt{wire\_cd} ($w$). The critical dimension of metal wire shapes, defined by wire width.
	\item[-] \texttt{track\_pitch} ($p$). The distance between adjacent wire tracks.
	\item[-] \texttt{min\_t2t} ($t_1$). The minimum line-end to line-end distance in each wire track. 
	\item[-] \texttt{max\_t2t} ($t_2$). The maximum line-end to line-end distance in each wire track. 
	\item[-] \texttt{min\_length} ($l_1$). The minimum length of a single wire shape along the wire tracks.
	\item[-] \texttt{max\_length} ($l_2$). The maximum length of a single wire shape along the wire tracks.
	\item[-] \texttt{t2t\_grid} ($t_g$). The unit size of line-end to line-end distance.
	\item[-] \texttt{total\_x} ($x_t$). The cell bounding box size in $x$ direction.
	\item[-] \texttt{total\_y} ($y_t$). The cell bounding box size in $y$ direction.
\end{itemize} 
To make a better understanding of these concepts, we also visualize certain terminologies in \Cref{fig:def-metal}.
It can be seen that \texttt{min\_t2t}, \texttt{max\_t2t}, \texttt{min\_length}, and \texttt{max\_length} make key contributions to metal distribution and density.

Metal shapes are generated track-by-track and each track is filled as described in \Cref{alg:otwg}.
We draw shapes from an initial lower-left coordinate $(x,y)$ (line 2), which
will be increased until $x$ reaches the total cell size constraint $x_t$ during the generation procedure (line 3);
the wire length is determined by a random number between $l_1$ and $\max(x_t-x,l_2)$ that ensures the shape length will never exceed the cell size (line 4);
we then draw a rectangle defined by the lower-left and upper-right coordinates (lines 5 -- 6);
spacing between adjacent shapes on single track is generated by line-end to line-end constraints $t_1$ and $t_2$ (line 7)
followed by the update of start coordinates (line 8).
The entire cell is filled track-by-track that can be totally in parallel as different tracks are generated independently, as in \Cref{alg:wcg}.

\begin{algorithm}[htb!]
	\caption{On Track Wire Generation}
	\label{alg:otwg}
	\begin{algorithmic}[1]
		\Function{\texttt{DrawWire}}{$w$, $t_1$, $t_2$, $t_g$, 
			$l_1$, $l_2$, $x_t$, $x_o$, $y_o$}
		\State Initialize parameters $x,y \gets 0$;
		\While{$x < x_t$}
		\State $l \gets \texttt{rand}(l_1, \max(l_2, x_t -x))$;
		\State $x_{ll} \gets x+x_o, y_{ll} \gets y+y_o, x_{ur} \gets x+l+x_o, y_{ur} \gets y+w+y_o$;
		\State Draw a rectangle defined by $x_{ll}, y_{ll}, x_{ur}$ and $y_{ur}$;
		\State $t \gets \texttt{rand}(t_1, \min(t_2, x-x_{ur}), t_g)$;
		\State $x \gets x + l + t$;
		\EndWhile
		\EndFunction
	\end{algorithmic}
\end{algorithm}

\begin{algorithm}[htb!]
	\caption{Wire Cell Generation}
	\label{alg:wcg}
	\begin{algorithmic}[1]
		\Function{DrawWireCell}{$w$, $p$, $t_1$, $t_2$, $t_g$, 
			$l_1$, $l_2$, $x_t$, $y_t$, $x_o$, $y_o$}
		\State Initialize parameters $y \gets 0$;
		\While{$y < y_t$}
		\State \texttt{DrawWire}($w$, $t_1$, $t_2$, $t_g$, $l_1$, $l_2$, $x_t$, $x_o$, $y$);
		\State $y \gets y_o + p +w$;
		\EndWhile
		\State \Return Cell;
		\EndFunction
	\end{algorithmic}
\end{algorithm}

\begin{table*}[tb!]
	\centering
	\caption{Metal layer generation specs ($\mu m$).}
	\label{tab:metal-spec}
	\begin{tabular}{c|ccccccccc}
		\toprule
		cellname & wire\_cd & track\_pitch & min\_t2t & max\_t2t & min\_length & max\_length & t2t\_grid & total\_x & total\_y \\ \midrule
        \texttt{test1}    & 0.016    & 0.032        & 0.012    & 0.2      & 0.044       & 0.1         & 0.005     & 100      & 100      \\
		\texttt{test2}    & 0.016    & 0.032        & 0.012    & 0.2      & 0.044       & 0.1         & 0.012     & 100      & 100      \\
		\texttt{test3}    & 0.016    & 0.032        & 0.012    & 0.4      & 0.044       & 0.5         & 0.005     & 100      & 100      \\
		\texttt{test4}    & 0.016    & 0.032        & 0.012    & 0.4      & 0.044       & 0.5         & 0.012     & 100      & 100      \\
		\texttt{test5}    & 0.016    & 0.032        & 0.012    & 0.6      & 0.044       & 0.5         & 0.036     & 100      & 100      \\
		\texttt{test6}    & 0.016    & 0.032        & 0.012    & 0.6      & 0.044       & 0.5         & 0.072     & 100      & 100      \\ \bottomrule
	\end{tabular}
\end{table*}
\subsection{Via Generation}
Vias, as conductive connections between different metal layers, are most likely to occur at regions where metals in adjacent layers are overlapped in the plane.
The basic idea of via generation is creating candidate via locations with horizontal and vertical metal shapes that can be generated with \Cref{alg:otwg}.
Thus, we have additional via constraints besides those control wire shape generation.
\begin{itemize}
	\item[-] \texttt{via\_x} ($v_x$). Via size along $x$ direction.
	\item[-] \texttt{via\_y} ($v_y$). Via size along $y$ direction.
	\item[-] \texttt{density} ($\rho$). The probability of a via appearing at a candidate via location.
	\item[-] \texttt{enclosure\_x} ($e_x$). The minimum horizontal distance from a via to metal line-end.
	\item[-] \texttt{enclosure\_y} ($e_y$). The minimum horizontal distance from a via to metal line-end.
	\item[-] \texttt{via\_pitch\_x} ($v_{px}$). The minimum center-to-center distance of two vias in the same tract in $x$ direction.
	\item[-] \texttt{via\_pitch\_x} ($v_{py}$). The minimum center-to-center distance of two vias in the same tract in $y$ direction.
\end{itemize}
where \texttt{via\_x} and \texttt{via\_y} are directly satisfied by controlling \texttt{wire\_cd}s of two metal layers,
\texttt{density} contributes to a threshold whether we should place a via at a candidate via location.
Challenging problem comes with satisfying enclosure and pitch constraints.

\begin{figure}[tb!]
	\centering
	\includegraphics[width=.38\textwidth]{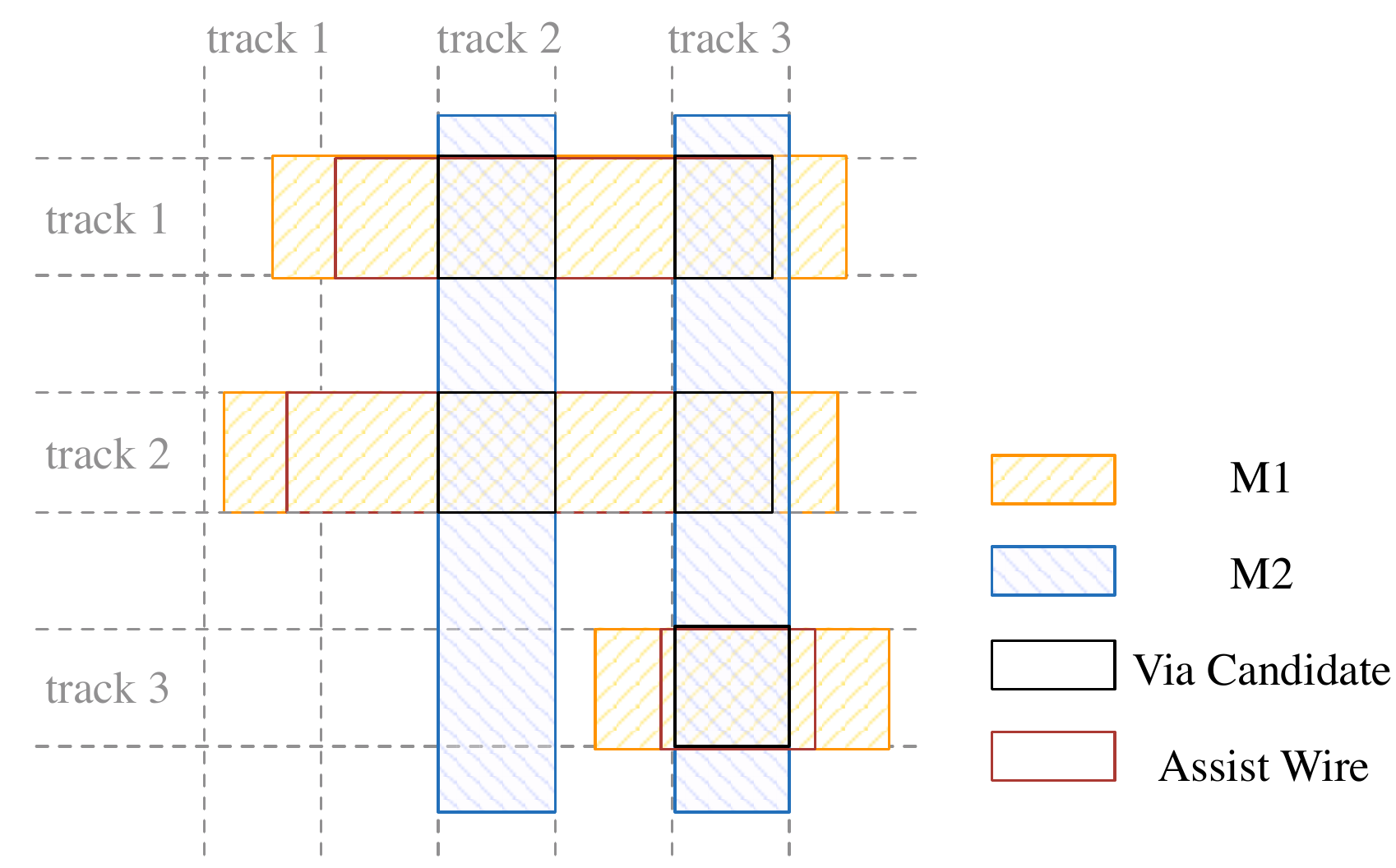}
	\caption{Satisfying enclosure constraints with assist wires.}
	\label{fig:assist-wire}
\end{figure}

We propose the concept of assist wire generation that generates on track wire shapes with line-ends shortened by the enclosure distance.
As shown in \Cref{fig:assist-wire}, we generate via candidate locations by taking the intersection between the assist layer and M2 layer,
instead of directly calculate the intersection between M1 and M2.
With the help of assist wire, we can direclty filter out vias that violate enclosure constraints by checking the area of them.
In real cases, enclosure constraints are applicable to both $x$ and $y$ directions,
which can be solved by simply using assist wires in vertical tracks.

To deal with the pitch constraints, we create a via candidate matrix for each cell, 
with each entry representing a cross point between a horizontal and a vertical tracks.
We put one at entries where there are metal shapes overlapping with each other and leave other entries zero.
The via candidate matrix for the cell in \Cref{fig:assist-wire} can then be written as follows.
\begin{align}
	\vec{M}_{vc}=\begin{bmatrix}
	0 &1 &\textcolor{red}{1} \\
	0 &1 &\textcolor{red}{1} \\
	0 &0 &1
	\end{bmatrix}.
\end{align}
After removing enclosure violations, the matrix becomes
\begin{align}
\vec{M}_{vc}=\begin{bmatrix}
0 &1 &\textcolor{red}{0} \\
0 &1 &\textcolor{red}{0} \\
0 &0 &1
\end{bmatrix}.
\end{align}
Because the metal pitch is intentionally picked the same as via during metal layer generation,
the problem of finding via pitch violations becomes finding $\begin{bmatrix}1&1\end{bmatrix}$ or $\begin{bmatrix}1\\1\end{bmatrix}$ patterns in $\vec{M}_{vc}$.
Finally, via pitch violations can be easily removed.

To summary, the via generation can be done with \Cref{alg:viag},
where we first generate metal layers that are to be connected by vias (lines 1--2);
candidate via shapes are placed by taking the intersection of two metal layers
along with the calculation of via candidate matrix (line 3);
following steps are removing enclosure conflicts (line 4), \texttt{pitch\_x} conflicts (lines 5--10) and \texttt{pitch\_y} conflicts (lines 11--16).

\begin{algorithm}[h]
	\caption{Via Generation}
	\label{alg:viag}
	\begin{algorithmic}[1]
		\Require M1 specs, M2 specs, $v_x$, $v_y$, $e_x$, $e_y$, $v_{px}$, $v_{py}$.
		\Ensure Via cell.
		\State $C_1 \gets$ Call \texttt{DrawWireCell} with M1 specs;
		\State $C_2 \gets$ Call \texttt{DrawWireCell} with M2 specs;
		\State $\vec{M}_{vc} \gets$ calculate via candidate matrix by taking $C_1$ \texttt{AND} $C_2$ and insert via shapes;
		\State $\vec{M}_{vc} \gets$ Detect enclosure conflicts and remove conflict vias;
		\For {each $\vec{M}_{vc}(i,j)$}
		\If{$\vec{M}_{vc}(i,j) = 1$ and $\vec{M}_{vc}(i-1,j) = 1$}
		\State Delete via cooresponding to $\vec{M}_{vc}(i,j)$;
		\State $\vec{M}_{vc}(i,j) \gets 0$;
		\EndIf
		\EndFor
		\For {each $\vec{M}_{vc}(i,j)$}
		\If{$\vec{M}_{vc}(i,j) = 1$ and $\vec{M}_{vc}(i,j-1) = 1$}
		\State Delete via cooresponding to $\vec{M}_{vc}(i,j)$;
		\State $\vec{M}_{vc}(i,j) \gets 0$;
		\EndIf
		\EndFor
	\end{algorithmic}
\end{algorithm}

\begin{table*}[tb!]
	\centering
	\caption{Via layer generation specs ($\mu m$).}
	\label{tab:via-spec}

	\begin{tabular}{c|ccccccccc}
		\toprule
		cellname & via1\_x & via1\_y & m1\_enc & m2\_enc & min\_via1\_pitch\_x & via\_fraction & min\_via1\_pitch\_y &total\_x &total\_y\\ \midrule
        \texttt{test1}    & 0.07    & 0.07    & 0.02    & 0.02    & 0.14                & 0.1           & 0.14     &100&100           \\
		\texttt{test2}    & 0.07    & 0.07    & 0.02    & 0.02    & 0.14                & 0.2           & 0.14      &100&100           \\
		\texttt{test3}    & 0.07    & 0.07    & 0.02    & 0.02    & 0.14                & 0.3           & 0.14     &100&100            \\
		\texttt{test4}    & 0.07    & 0.07    & 0.02    & 0.02    & 0.14                & 0.4           & 0.14     &100&100            \\
		\texttt{test5}    & 0.07    & 0.07    & 0.02    & 0.02    & 0.14                & 0.5           & 0.14    &100&100             \\
		\texttt{test6}    & 0.07    & 0.07    & 0.02    & 0.02    & 0.14                & 0.6           & 0.14    &100&100             \\ \bottomrule
	\end{tabular}
\end{table*}
\subsection{Layout Generation Performance}
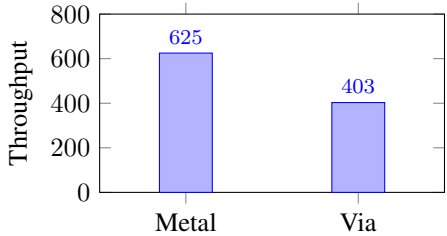
\begin{figure}
	\centering
	\begin{tikzpicture}
    \pgfplotsset{
        width =0.34\textwidth,
        height=0.218\textwidth,
        every axis plot/.append style = {font = \footnotesize}
    }
    \begin{axis}[
            ybar=0.5pt,
            ymin=0,
            ymax=800,
            enlarge x limits=0.5,
            bar width=20pt,
            legend style={at={(0.83,0.9)},
            anchor=north,legend columns=1},
            area legend,
            ylabel={Throughput},
            symbolic x coords={Metal, Via},
            xtick=data,
            ytick={0,200,400,600,800},
            ylabel near ticks,
            nodes near coords,
        ]
        \addplot  coordinates {(Metal,625) (Via,403)};

    \end{axis}
\end{tikzpicture}
    \caption{Throughput ($\mu m^2$ / s) estimation of the layout generation tool.}
	\label{fig:tp}
\end{figure}

We evalutate the performance of the proposed layout generation tool on an Intel Xeon 3.5GHz platform with single thread.
We adopt the specs in \Cref{tab:metal-spec} and \Cref{tab:via-spec} for metal and via layer generation, respectively,
with each test cell has an area of 0.001$mm^2$.
The throughputs are dipicted in \Cref{fig:tp} that corresponds to the area of metel/via cell generated per second.

\section{Evaluation of Representative Hotspot Detectors}
\label{sec:exp}
As a case study, we generate a group of via patterns with above via generation flow and obtain clip labels by feeding them into an industrial DRC and simulation tool.
In this section, we will use the generated dataset to evaluate the performance of different loss functions on several representative machine learning-based hotspot detectors \cite{HSD-TCAD2019-Yang,HSD-ICCAD2016-Zhang,HSD-ASPDAC2019-Ye}. 
We first introduce beneath ideas of these hotspot detectors.

\subsection{Feature Tensor Extraction and Batch Biased Learning (BBL) \cite{HSD-TCAD2019-Yang}}
Layout hotspot detection, originally, can be regarded as an image classification problem.
However, the input size is much larger than images in traditional classification tasks because a large area of layout context is required to determine whether a clip contains hotspot or not.
\cite{HSD-TCAD2019-Yang} tackles the problem with a layout compression technique in frequency domain that dramatically reduces input size as well as preserve a large fraction of original layout information.
The compression is conducted following four steps including (1) layout slicing based on critical dimension (2) DCT on sub clip blocks (3) flatten DCT coefficients in zig-zag form and (4) discarding high frequency components and constructing feature tensor.
Additionally, the compressed data can be easily learned with shallow neural networks. 
As a case study, we will use the architecture specified in \Cref{tab:arch} in all the deep learning related experiments.
\begin{table}[tb!]
    \centering
    \caption{Neural Network Configuration.}
    \label{tab:arch}
    \small
    \begin{tabular}{c|c|c|c}
        \toprule 
        Layer     & Kernel Size & Stride   &Output Node \# \\
        \midrule
        conv1-1       &3     &1      &$12 \times 12 \times 16$  \\
        conv1-2       &3     &1      &$12 \times 12 \times 16$  \\
        maxpooling1   &2     &2      &$6 \times 6 \times 16$    \\
        \midrule
        conv2-1       &3     &1      &$6 \times 6 \times 32$    \\
        conv2-2       &3     &1      &$6 \times 6 \times 32$    \\
        maxpooling2   &2     &2      &$3 \times 3 \times 32$    \\
        \midrule
        fc1           &-     &-      &250  \\
        fc2           &-     &-      &2    \\
        \bottomrule 
    \end{tabular}
\end{table}

In the procedure of training a hotspot detector, neuron weights are updated towards pattern labels. 
\cite{HSD-TCAD2019-Yang} takes advantage of the fact that any non-hotspot instances with predicted non-hotspot probability over 50\% will be correctly classified.
Therefore, it is not necessary to guide the neural networks to learn a high confidence prediction.
Accordingly, a batch biased learning is proposed to introduce label penalty on non-hotspot patterns based on their training loss on current step which in turn controls the training label penalty.
\subsection{Smooth Boosting \cite{HSD-ICCAD2016-Zhang}}
\cite{HSD-ICCAD2016-Zhang} provides a legacy machine learning solution for hotspot detection which includes feature engineering and learning engine development.
In the feature extraction stage, \cite{HSD-ICCAD2016-Zhang} improves traditional concentric circle area sampling (CCAS) \cite{HSD-SPIE2015-Matsunawa} with mutual information, which considers that only partial of sampled circles have strong contributions to final feature vectors and labels.
Circle selection problem can be formulated as follows.
\begin{subequations}
\begin{align}
    \mathcal{I}^\ast_{n_c} = &\arg \min_{\mathcal{I}_{n_c}\subseteq \mathcal{I}} \sum_{i \in \mathcal{I}_{n_c}} I(C_i;Y), \\
    &\text{s.t.~}|i-j|>d,\forall i,j\in \mathcal{I}_{n_c},
\end{align}
\end{subequations}
where $\mathcal{I}^\ast_{n_c}$ includes selected circle indices, $\mathcal{I}=\{i|1<i<r_\max, i\in \mathbb{N}\}$ contains indices of all circle candidates and
\begin{align}
    I(C_i;Y) = \sum_{c_i \in C_i} \sum_{y\in Y} p(c_i,y)\log \dfrac{p(c_i,y)}{p(c_i)p(y)},
\end{align}
which calculates the mutual information of $i_{th}$ circle and labels.
Here $C_i$s and $Y$ are random variables defined in circle index space and label space.

Smooth Boosting \cite{servedio2003smooth}, as an ensemble learning engine with good noise tolerance, is chosen as the preferred model here for hotspot detection tasks,
where the weight of each weak classifier is updated during training such that each weak classifier satisfies \Cref{eq:wc}.
\begin{align}
\label{eq:wc}
\frac{1}{2} \sum_{j=1}^{n} M_{t}(j)\left|h_{t}\left(\mathbf{x}_{j}\right)-y_{j}\right| \leq \frac{1}{2}-\gamma,
\end{align}
where $M_{t}(j)$s are weight terms, $h_t$s are hypothesis function of each weak classifier and $\gamma$ is a hyper-parameter defined in \cite{servedio2003smooth}.

In the inference stage, all weak learners will work together to predict the label of input clips with their weighted output being the final classification results, as in \Cref{eq:inf}.
\begin{align}
\label{eq:inf}
    f = \operatorname{sign}\left(\frac{1}{T} \sum_{t=1}^{T} h_{t}\right).
\end{align}

\subsection{LithoROC \cite{HSD-ASPDAC2019-Ye}}
Receiver operating characteristic (ROC) has be widely used as a machine learning model evaluation metric.
Given a dataset $D=\{(x_i, y_i)\}_{i=1}^N$, where $x_i \in {\mathbb{R}}^d$ and $y_i \in \{-1,+1\}$ are the $i$-th data sample in the feature space and its true class label, we can divide the dataset into a positive sample set $D_+=\{(x_i^+,+1\}_{i=1}^{N^+}$ and a negative sample set $D_-=\{(x_i^-,-1\}_{i=1}^{N^-}$, where $N_+$ and $N_-$ denote the number of positive and negative samples respectively, and $N={N_+}+{N_-}$. Let $f(x)$ denote the prediction model.

Based on the equivalence of area under the ROC curve(AUC) and Wilcoxon-Mann-Whitney (WMW) statistic test of ranks, pairwise convex surrogate loss $\Phi(f(x_i^+)-f(x_j^-))$ is applied to make the AUC differentiable. The original AUC function is defined as following:
\begin{equation}
    {AUC= \frac{1}{N_+ N_-} \sum_{i=1}^{N_+}\sum_{j=1}^{N_-}I(f(x_i^+)-f(x_j^-))},
\end{equation}
where $I((f(x_i^+)-f(x_j^-)))$ is the discontinuous indicator function. 
The differentiable form is shown below to work as the loss function.
\begin{equation}
L_\Phi(f)= \frac{1}{N_+ N_-} \sum_{i=1}^{N_+}\sum_{j=1}^{N_-}\Phi(f(x_i^+)-f(x_j^-)).
\end{equation}
Then maximizing the AUC score is equivalent to minimizing the loss function $L_\Phi(f)$.
Let $z=f(x_i^+)-f(x_j^-)$, \cite{HSD-ASPDAC2019-Ye} recommend four surrogate loss functions for differentiable AUC approximation:
the pairwise squared loss (PSL) \cite{gao2013one}, the pairwise hinge loss (PHL) \cite{steck2007hinge}, the pairwise logistic loss (PLL) \cite{rudin2009margin},
and the R loss function \cite{yan2003optimizing} as shown below.
\begin{align}
    \label{eq:psl}
    &\Phi_{PSL}(z) =(1-z)^2, \\
    &\Phi_{PHL}(z) = \max(1-z, 0), \\
    &\Phi_{PLL}(z) = \log (1 + \exp(-\beta z)), \\
    \label{eq:r}
    &\Phi_{R}(z) =     \left\{
        \begin{aligned}
            & {(-(z-\gamma))^p},&&\text{if $z>\gamma$}, \\
            & 0,                &&\text{otherwise}. 
        \end{aligned}
        \right.
\end{align}
It should also be noted that in R loss function we have $0< \gamma <1$ and $p>1$.
Additionally, we heuristically define two new cubic loss functions (Pairwise Cubic Loss Function1 and Pairwise Cubic Loss Function2) and test their performances and compare with previous ones.
\begin{align}
   \Phi_{PCL1}(z) &= \max(8 - (1+z)^3,0),\\
   \Phi_{PCL2}(z) &= \max((1-z)^3,0).
\end{align}

\begin{figure}
    \centering
    \includegraphics[width=.86\linewidth]{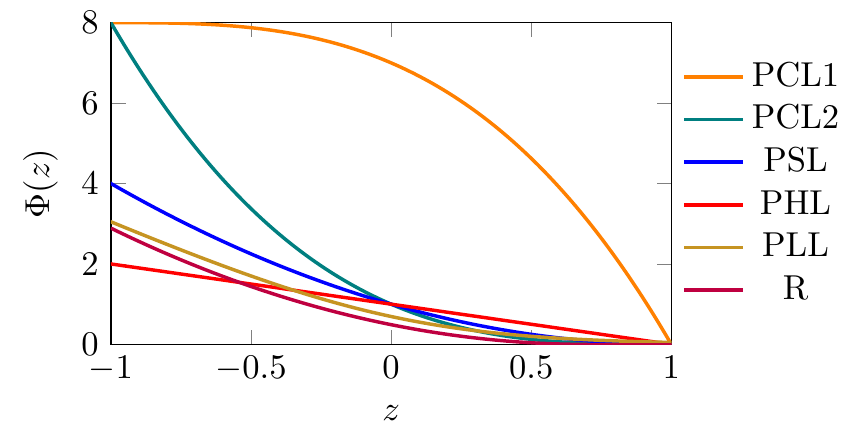}
    \caption{Visualization of different approximations of AUC.}
    \label{fig:auc}
\end{figure}

\Cref{fig:auc} illustrates the comparison of all surrogate loss functions. Note that there is a sharp increase in penalty when $z$ goes to -1 in our cubic functions. The purpose is to apply more penalty when false positive and false negative simultaneously appear. As we can see from the curve of PCL2, the penalty for $z$ is highly to the original ones when $z \in [0,1]$. 
Furthermore, we can also clearly distinguish larger penalty when any mis-prediction occurs (for curve region $[-1,0]$), which is expected to result in slightly better performance.

\subsection{Experiments}
\begin{table*}[htb!]
    \centering
    \caption{CNN-based hotspot detectors behave unstably on challenging datasets.}
    \label{tab:exp-res}
    \setlength{\tabcolsep}{3pt}
    \begin{tabular}{c|cc|cc|cc|cc|cc|cc|cc}
        \midrule
        \multirow{2}{*}{ID} & \multicolumn{2}{c|}{PSL} & \multicolumn{2}{c|}{PHL} & \multicolumn{2}{c|}{PLL} & \multicolumn{2}{c|}{R} & \multicolumn{2}{c|}{PCL1} & \multicolumn{2}{c|}{PCL2} & \multicolumn{2}{c}{BBL}\\ \cmidrule{2-15}
        & Acc (\%)& FA (\%) & Acc (\%) & FA (\%) & Acc (\%) & FA (\%) & Acc (\%) & FA (\%) & Acc (\%)  & FA (\%) & Acc (\%) & FA (\%)& Acc (\%) & FA (\%)\\
        \midrule
        1    & 10.7    & 0.96       & 94.17    & 60.95     & 87.87      & 43.79    & 10.82    & 1.92   & 81.33    & 36.36  &24.14    & 2.82    &58.74    &24.50\\
        2    & 17.48   & 2.5        & 79.31    & 38.28     & 53.26      & 16.90    &30.56     & 7.81   &83.59     &39.37   &86.09    & 44.69   &62.78    &23.43\\
        3    & 17.6    & 3.26       & 93.34    & 59.54     & 33.17      & 6.59     &17.95     & 2.11   &40.07     & 8.77   &85.25    & 40.90   &60.64    &22.92\\
        4    & 20.81   & 4.23       & 38.76    & 7.94      & 28.18      & 4.99     &30.92     &7.11    &92.39     &51.34   &72.53    &27.66    &65.40    &31.37\\
        5    & 18.67   & 3.07       & 17.36    & 1.66      & 64.09      & 22.86    &15.57     &1.86    &85.49     &39.88   &46.25    &11.46    &58.26    &24.71\\ \midrule
        Ave  & 17.05   & 2.80       & 64.59    & 33.67     & 53.31      & 19.03    &21.16     &4.16    &76.57     &35.14   &62.85    &25.51    &61.16    &25.39\\ 
        Var  & 14.39   & 1.45       & 1204.02  & 780.35    & 586.94     & 246.03   &83.02     &9.13    &433.51    &249.92  &727.40   &330.40   &8.78     &11.74\\
        \bottomrule
        \hline
    \end{tabular}
\end{table*}
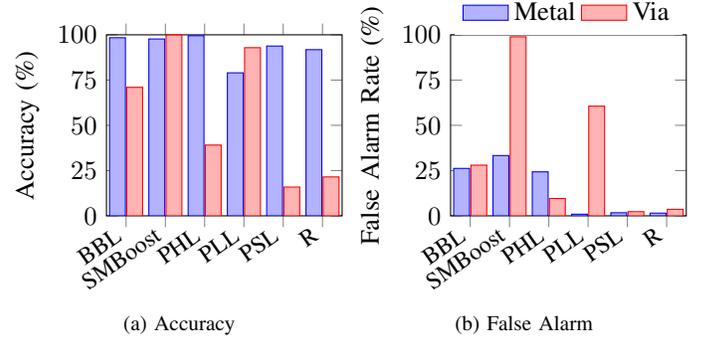
\begin{figure}[tb!]
    \centering
    \subfloat[Accuracy]{\begin{tikzpicture}
\pgfplotsset{
	width =0.26\textwidth,
	height=0.22\textwidth,
	every axis plot/.append style = {font = \tiny}
}
\begin{axis}[
ybar=0.5pt,
ymin=0,
ymax=100,
enlarge x limits=0.1,
bar width=6pt,
legend style={at={(0.5,1.0)},
	anchor=north,legend columns=-1},
area legend,
ylabel={Accuracy (\%)},
symbolic x coords={BBL, SMBoost, PHL, PLL, PSL, R},
xtick=data,
ytick={0,25,50,75,100},
ylabel near ticks,
x tick label style={rotate=35,anchor=east,font=\small},
]
\addplot  coordinates {(BBL,98.4) (SMBoost,97.7)(PHL,99.5) (PLL,79.0) (PSL,93.8) (R,91.8)};
\addplot  coordinates {(BBL,71) (SMBoost,99.9)(PHL,39.1) (PLL,92.9) (PSL,15.9) (R,21.5)};

\end{axis}
\end{tikzpicture}}
    \subfloat[False Alarm]{\begin{tikzpicture}
\pgfplotsset{
	width =0.26\textwidth,
	height=0.22\textwidth,
	every axis plot/.append style = {font = \tiny}
}
\begin{axis}[
ybar=0.5pt,
ymin=0,
ymax=100,
enlarge x limits=0.1,
bar width=6pt,
legend style={
    draw=none,
    at={(0.5,1.0)},
	anchor=south,legend columns=2},
area legend,
ylabel={False Alarm Rate (\%)},
symbolic x coords={BBL, SMBoost, PHL, PLL, PSL, R},
xtick=data,
ytick={0,25,50,75,100},
ylabel near ticks,
x tick label style={rotate=35,anchor=east,font=\small},
]
\addplot  coordinates {(BBL,26.2) (SMBoost,33.3)(PHL,24.3) (PLL,0.9) (PSL,1.7) (R,1.5)};
\addplot  coordinates {(BBL,28) (SMBoost,98.9)(PHL,9.6)(PLL,60.6)(PSL,2.3)(R,3.6)};

\legend{Metal, Via}
\end{axis}
\end{tikzpicture}}
    \caption{Evaluation of machine learning-based hotspot detectors.}
    \label{fig:rst}
\end{figure}
We implement these hotspot detection solutions with $\mathsf{Python}$ and $\mathsf{Tensorflow}$ 1.9.
All experiments are conducted on a Intel platform with Titan Xp graphic unit.
For the experiments on BBL and SMBoost, we adopt the same parameter settings as in \cite{HSD-TCAD2019-Yang} and \cite{HSD-ICCAD2016-Zhang} respectively.
In the experiments on LithoRoC, we explore the performances of \cite{HSD-TCAD2019-Yang} with surrogate loss function suggested by \cite{HSD-ASPDAC2019-Ye} to seek for the increasing and stabilizing the hotspot detection accuracy and minimizing false alarms. In our experiment, we adopt the original unified data from \cite{HSD-TCAD2019-Yang}. The detector is evaluated with $\gamma=1\mathrm{e}-3$, $\alpha=0.65$, $m=32$ and $k=2000$. We set $\beta=3$ in PLL loss function and $\gamma=0.7$, $p=2$ in R loss function. In each iteration, $\frac{m}{2}$ hotspot and $\frac{m}{2}$ non-hotspot training instances are randomly selected to perform weight updating. Learning rate decays after every $k$ iterations. During optimization, we randomly select one between-class pair to calculate $z$. After each iteration, a new batch is generated, so all between-class pairs have equal probability to be fetched for optimization. In the end, we use the last saved model to test prediction performance,
which is measured following previous works \cite{HSD-ICCAD2012-Torres}.
\begin{mydefinition}[Accuracy (Acc)]
	\label{def:acc}
	The ratio between the number of correctly predicted hotspot clips and the number of all real hotspot clips.
\end{mydefinition}
\begin{mydefinition}[False Alarm Rate (FA)]
	\label{def:fa}
	The ratio between the number of correctly predict non-hotspot clips and the number of all non-hotspot clips.
\end{mydefinition}

To explore the robustness of the network, we perform five parallel experiments and differentiate them with IDs, as shown in the first column of \Cref{tab:exp-res}. Adjacent columns display detailed results with headers ``PSL'', ``PHL'', ``PLL'' and ``R'',
which correspond to the AUC approximation function defined from \Cref{eq:psl} to \Cref{eq:r} respectively.
Columns ``Acc'' and ``FA'' denote hotspot detection accuracy and false alarm in the percentage form respectively. 
In the last two rows, we compute the average and variance after Bessel's Correction
for each column to estimate their performance and stability.
It can be seen that although these training solutions behave reasonably good on some designs (see blue bars of \Cref{fig:rst}), they all fail on challenging datasets.
Furthermore, some of the existing solutions rely highly on the initial status of the neural networks and hence yield unstable performance, as listed in \Cref{tab:exp-res}.


\section{Conclusion}
\label{sec:conclu}
In this paper, we propose a test layout generation toolkit built upon $\mathsf{KLayout}$ backbone.
The toolkit supports both contact/via and uni-directional metal layer generation with certain design rule configurations.
The tool can also be easily extended to other purposes including cell processing, complicated test pattern generation (e.g.~Hilbert space patterns) and layout transformation.
As a case study, we generate a set of DRC-clean via layout clips to evaluate the robustness and stability of state-of-the-art machine learning-based hotspot detectors.
Here we introduce three recent hotspot detectors that adopt batch biased learning, smooth boosting and AUC optimization respectively,
which all exhibit good performance on public metal layer design but unstable on the generated via designs.
The study shows continuous research are necessary to increase the robustness and stability of machine learning models and hence prototype such frameworks into real back-end design flow.

\section*{Acknowledgments}
This work is supported in part by The Research Grants Council of Hong Kong SAR (Project No.~CUHK24209017).

\bibliographystyle{IEEEtran}
\bibliography{./ref/Top-sim,./ref/HSD,./ref/DFM,./ref/Additional}

\begin{thebibliography}{10}
\providecommand{\url}[1]{#1}
\csname url@samestyle\endcsname
\providecommand{\newblock}{\relax}
\providecommand{\bibinfo}[2]{#2}
\providecommand{\BIBentrySTDinterwordspacing}{\spaceskip=0pt\relax}
\providecommand{\BIBentryALTinterwordstretchfactor}{4}
\providecommand{\BIBentryALTinterwordspacing}{\spaceskip=\fontdimen2\font plus
\BIBentryALTinterwordstretchfactor\fontdimen3\font minus
  \fontdimen4\font\relax}
\providecommand{\BIBforeignlanguage}[2]{{%
\expandafter\ifx\csname l@#1\endcsname\relax
\typeout{** WARNING: IEEEtran.bst: No hyphenation pattern has been}%
\typeout{** loaded for the language `#1'. Using the pattern for}%
\typeout{** the default language instead.}%
\else
\language=\csname l@#1\endcsname
\fi
#2}}
\providecommand{\BIBdecl}{\relax}
\BIBdecl

\bibitem{HSD-ICCAD2016-Zhang}
H.~Zhang, B.~Yu, and E.~F.~Y. Young, ``Enabling online learning in lithography
  hotspot detection with information-theoretic feature optimization,'' in
  \emph{Proc.~ICCAD}, 2016, pp. 47:1--47:8.

\bibitem{HSD-ASPDAC2019-Yang}
H.~Yang, P.~Pathak, F.~Gennari, Y.-C. Lai, and B.~Yu, ``Detecting multi-layer
  layout hotspots with adaptive squish patterns,'' in \emph{Proc.~ASPDAC},
  2019, pp. 299--304.

\bibitem{HSD-SPIE2015-Matsunawa}
T.~Matsunawa, J.-R. Gao, B.~Yu, and D.~Z. Pan, ``A new lithography hotspot
  detection framework based on {AdaBoost} classifier and simplified feature
  extraction,'' in \emph{Proc.~SPIE}, vol. 9427, 2015.

\bibitem{HSD-TCAD2017-Yang}
F.~Yang, S.~Sinha, C.~C. Chiang, X.~Zeng, and D.~Zhou, ``Improved tangent space
  based distance metric for lithographic hotspot classification,'' \emph{IEEE
  TCAD}, vol.~36, no.~9, pp. 1545--1556, 2017.

\bibitem{HSD-ASPDAC2017-Tomioka}
Y.~Tomioka, T.~Matsunawa, C.~Kodama, and S.~Nojima, ``Lithography hotspot
  detection by two-stage cascade classifier using histogram of oriented light
  propagation,'' in \emph{Proc.~ASPDAC}, 2017, pp. 81--86.

\bibitem{HSD-ISVLSI2018-Geng}
H.~Geng, H.~Yang, B.~Yu, X.~Li, and X.~Zeng, ``Sparse {VLSI} layout feature
  extraction: A dictionary learning approach,'' in \emph{Proc.~ISVLSI}, 2018,
  pp. 488--493.

\bibitem{HSD-SOCC2017-Yang}
H.~Yang, Y.~Lin, B.~Yu, and E.~F.~Y. Young, ``Lithography hotspot detection:
  From shallow to deep learning,'' in \emph{Proc.~SOCC}, 2017, pp. 233--238.

\bibitem{HSD-TCAD2019-Yang}
H.~Yang, J.~Su, Y.~Zou, Y.~Ma, B.~Yu, and E.~F.~Y. Young, ``Layout hotspot
  detection with feature tensor generation and deep biased learning,''
  \emph{IEEE TCAD}, vol.~38, no.~6, pp. 1175--1187, 2019.

\bibitem{HSD-DAC2019-Chen}
R.~Chen, W.~Zhong, H.~Yang, H.~Geng, X.~Zeng, and B.~Yu, ``Faster region-based
  hotspot detection,'' in \emph{Proc.~DAC}, 2019, pp. 146:1--146:6.

\bibitem{HSD-ASPDAC2019-Ye}
W.~Ye, Y.~Lin, M.~Li, Q.~Liu, and D.~Z. Pan, ``{LithoROC}: lithography hotspot
  detection with explicit {ROC} optimization,'' in \emph{Proc.~ASPDAC}, 2019,
  pp. 292--298.

\bibitem{HSD-ASPDAC2012-Ding}
D.~Ding, B.~Yu, J.~Ghosh, and D.~Z. Pan, ``{EPIC}: Efficient prediction of {IC}
  manufacturing hotspots with a unified meta-classification formulation,'' in
  \emph{Proc.~ASPDAC}, 2012, pp. 263--270.

\bibitem{HSD-JM3-2017-Yang}
H.~Yang, L.~Luo, J.~Su, C.~Lin, and B.~Yu, ``Imbalance aware lithography
  hotspot detection: a deep learning approach,'' \emph{JM3}, vol.~16, no.~3, p.
  033504, 2017.

\bibitem{HSD-JM3-2016-Shin}
M.~Shin and J.-H. Lee, ``Accurate lithography hotspot detection using deep
  convolutional neural networks,'' \emph{JM3}, vol.~15, no.~4, p. 043507, 2016.

\bibitem{HSD-ICCAD2012-Torres}
A.~J. Torres, ``{ICCAD-2012 CAD} contest in fuzzy pattern matching for physical
  verification and benchmark suite,'' in \emph{Proc.~ICCAD}, 2012, pp.
  349--350.

\bibitem{HSD-DAC2019-DeePattern}
H.~Yang, P.~Pathak, F.~Gennari, Y.-C. Lai, and B.~Yu, ``{DeePattern}: Layout
  pattern generation with transforming convolutional auto-encoder,'' in
  \emph{Proc.~DAC}, 2019, pp. 148:1--148:6.

\bibitem{HSD-JM3-2015-Shim}
S.~Shim and Y.~Shin, ``Topology-oriented pattern extraction and classification
  for synthesizing lithography test patterns,'' \emph{JM3}, vol.~14, no.~1, pp.
  013\,503--013\,503, 2015.

\bibitem{KLAYOUT}
``{KLAYOUT},'' \url{https://www.klayout.de}.

\bibitem{servedio2003smooth}
R.~A. Servedio, ``Smooth boosting and learning with malicious noise,''
  \emph{Journal of Machine Learning Research}, vol.~4, no. Sep, pp. 633--648,
  2003.

\bibitem{gao2013one}
W.~Gao, R.~Jin, S.~Zhu, and Z.-H. Zhou, ``One-pass auc optimization,'' in
  \emph{Proc.~ICML}, 2013, pp. 906--914.

\bibitem{steck2007hinge}
H.~Steck, ``Hinge rank loss and the area under the roc curve,'' in
  \emph{European Conference on Machine Learning}.\hskip 1em plus 0.5em minus
  0.4em\relax Springer, 2007, pp. 347--358.

\bibitem{rudin2009margin}
C.~Rudin and R.~E. Schapire, ``Margin-based ranking and an equivalence between
  adaboost and rankboost,'' \emph{Journal of Machine Learning Research},
  vol.~10, no. Oct, pp. 2193--2232, 2009.

\bibitem{yan2003optimizing}
L.~Yan, R.~H. Dodier, M.~Mozer, and R.~H. Wolniewicz, ``Optimizing classifier
  performance via an approximation to the wilcoxon-mann-whitney statistic,'' in
  \emph{Proc.~ICML}, 2003, pp. 848--855.

\end{thebibliography}

\end{document}